\documentclass[conference]{IEEEtran}
\IEEEoverridecommandlockouts
\usepackage{cite}
\usepackage{amsmath,amssymb,amsfonts}
\usepackage{algorithmic}
\usepackage{graphicx}
\usepackage{textcomp}
\usepackage{xcolor}
\usepackage{bm}
\usepackage{subcaption}
\usepackage{times}  
\usepackage{helvet}  
\usepackage{courier}  
\usepackage{url}  

\usepackage{color}

\def\BibTeX{{\rm B\kern-.05em{\sc i\kern-.025em b}\kern-.08em
    T\kern-.1667em\lower.7ex\hbox{E}\kern-.125emX}}
\begin{document}

\title{A Deep Structural Model for Analyzing Correlated Multivariate Time Series}
\author{
\IEEEauthorblockN{Changwei Hu}
\IEEEauthorblockA{\textit{Yahoo Research} \\
New York, USA \\
ch237duke@gmail.com}
\and
\IEEEauthorblockN{Yifan Hu}
\IEEEauthorblockA{\textit{Yahoo Research} \\
	New York, USA \\
	yifanhu@verizonmedia.com}
\and
\IEEEauthorblockN{Sungyong Seo}
\IEEEauthorblockA{\textit{University of Southern California} \\
	Los Angeles, USA \\
	sungyons@usc.edu}
}
\maketitle

\begin{abstract}
Multivariate time series are routinely encountered in real-world applications, and in many cases, these time series are strongly correlated. In this paper, we present a deep learning structural time series model which can (i) handle correlated multivariate time series input, and (ii) forecast the targeted temporal sequence by explicitly learning/extracting the trend, seasonality, and event components. The trend is learned via a 1D and 2D temporal CNN and LSTM hierarchical neural net. The CNN-LSTM architecture can (i) seamlessly leverage the dependency among multiple correlated time series in a natural way, (ii) extract the \emph{weighted differencing} feature for better trend learning, and (iii) memorize the long-term sequential pattern. The seasonality component is approximated via a non-liner function of a set of Fourier terms, and the event components are learned by a simple linear function of regressor encoding the event dates. We compare our model with several state-of-the-art methods through a comprehensive set of experiments on a variety of time series data sets, such as forecasts of Amazon AWS Simple Storage Service (S3) and Elastic Compute Cloud (EC2) billings, and the closing prices for corporate stocks in the same category.
\end{abstract}


\section{Introduction}
Accurate forecasting of time series is a fundamental challenge for machine learning and it naturally requires understanding the sequential behavior. Traditionally, many studies on time series have focused on learning \emph{internal patterns} (e.g., auto-correlation) in a given series. However, most of the time series in many practical applications are highly correlated to other time series. For instance, a stock price of a particular company is likely dependent on the prices of other companies in similar business field~\cite{pai2005hybrid,lahmiri2018minute}. Since many multivariate time series are mutually related, it is desired to learn not only intra-series patterns but also inter-series patterns. Moreover, the time series we observe have shown seasonality or abrupt rising/dropping patterns due to specific events. Recently, many works~\cite{sean2018,kay2015} have analyzed a given series as interactions of different components such as the trend, seasonality, and event. While such an analysis is interpretable and mostly works well on many time series, a key open challenge is developing methods that can directly learn entangled relations among multiple time series.
In other words, it is beneficial to build a model to handle both characteristics, 1) highly correlated multivariate and 2) multiple structural components, such as trend, seasonality, and events, for accurate time series forecasts.

In this paper, we develop a framework to model correlated and structural time series. Inspired by the recent work of using 1D CNN for text classification ~\cite{kim2014}, we propose to use 1D CNN layer with multiple kernels to learn the complicated interactions among multivariate time series. However, unlike 1D CNN text classification, our model performs the 1D convolution over the feature dimension, rather than the temporal dimension. To improve the forecasts of time series, especially the forecasts of the trend, we further propose to use a 2D convolutional filters over the temporal dimension. In computer vision, convolutional filters are applied to local area of a given image, which aim to extract features corresponding to the structure of the filters. For example, if the filter is $(-1,1)$, the convolution computes the \emph{difference of adjacent pixels' values} and can help to detect an edge feature. These properties are equivalently applicable to time series and the gradient-based filters can be very useful to model a function based on first order derivative of each series. This is somewhat similar in spirit to the first-order differencing, a technique frequently used by time series forecasting models, such as ARIMA, to extract/eliminate the trend in the temporal sequences.

Recently, many deep learning models are data driven, instead of being based on carefully hand-crafted architectures. 
Among them, recurrent neural networks (RNNs) including LSTM~\cite{hochreiter1997long} and GRU~\cite{cho2014properties} are especially designed to learn sequential dependencies. These models are powerful to learn long term patterns by reducing the troublesome vanishing gradients~\cite{hochreiter1998vanishing}. To memorize the long-term sequential pattern while leveraging the
correlations among different time series, we stack a LSTM layer on top of our 1D and 2D CNN layers. Since the features extracted by the 1D and 2D convolutional filters can be stacked along the temporal dimension,
these extracted features are naturally time series, and thus can be fed into LSTMs to learn the long range temporal pattern changes in trend. The combination of CNN and RNN has been studied in video recognition~\cite{donahue2015long} and natural language model~\cite{kim2016character}. \cite{donahue2015long} used convolutional filters to extract features from still images and did not consider any differencing features between adjacent images.
\cite{kim2016character} has multiple character-level convolutional filters to extract character $n$-gram features, however, they do not use 1D convolutional filters to extract correlation features.

Besides the trend component learned by the convolutional filters, the seasonal and the event-based components are considered as additional components that compose the time series. These components have unique properties such as periodic behavior and discontinuous peaks, respectively.
These properties give us a prior knowledge to build an expressive model. For instance, a known periodical function (e.g., sinusoidal functions) can be a base function to represent the cyclic characteristics efficiently.
For the event-based components, a binary vector encoding is used to tell a model when the events happen.
In our model, the weights of both components would be updated by data-driven learning.

To summarise, our paper has the following contributions:
\begin{itemize}
	\item Our model employs 1D CNN and 2D CNN to learn multivariate correlations and gradient-based features (weighted differencing). By stacking LSTM layer on top of the CNNs, our proposed model is able to learn long-term dependencies (e.g., trend) of the extracted features.
	\item Our structural model is capable of decomposing time series into multiple latent structural factors including trend, seasonality, and event components, which is helpful for the interpretation of the time series forecasts.
	\item Our model is compared with several state-of-the-art baselines on Amazon AWS daily billing data for different services and the closing price for stocks, and shows improvement in forecasting accuracy.
\end{itemize}

\section{The Model}

In this section, we will introduce our proposed model by starting from structural time series (STM) analysis, a framework based on which our model is built. Then, we will describe problem formulation and each component of our model in more details.

\subsection{Structural Time Series Analysis}
The STM framework assumes that a temporal sequence is generated by several independent additive components, which have a direct interpretation in terms of quantities of interest. There exist many variations of STM models, depending how each component is defined. Here we are interested in a specific STM model~\cite{harvey1990}, in which a time series is decomposed into three components: the trend, seasonality, and event-related item. Denote the time series at time $t$ as $\bm{x}_t$, the model can be formulated as
\begin{equation}
\vspace{-6pt}
\bm{x}_t = \bm{d}_t + \bm{s}_t +\bm{e}_t
\label{eq:stm}
\vspace{-4pt}
\end{equation}
where $\bm{d}_t$, $\bm{s}_t$, and $\bm{e}_t$ are trend term, seasonality term, and event term, respectively. The STM model enables data analysts to analyze the trend, seasonal patterns, and irregular event effects of the data. It also allows for incorporation of prior knowledge such as seasonality, irregular event effects into time series forecasting. Many time series models, such as Prophet~\cite{sean2018} and Bayesian STM~\cite{kay2015}, fall into the category of this formulation.
\begin{figure}[h]
	\vspace{-12pt}
	\centering
	\includegraphics[scale = 0.4]{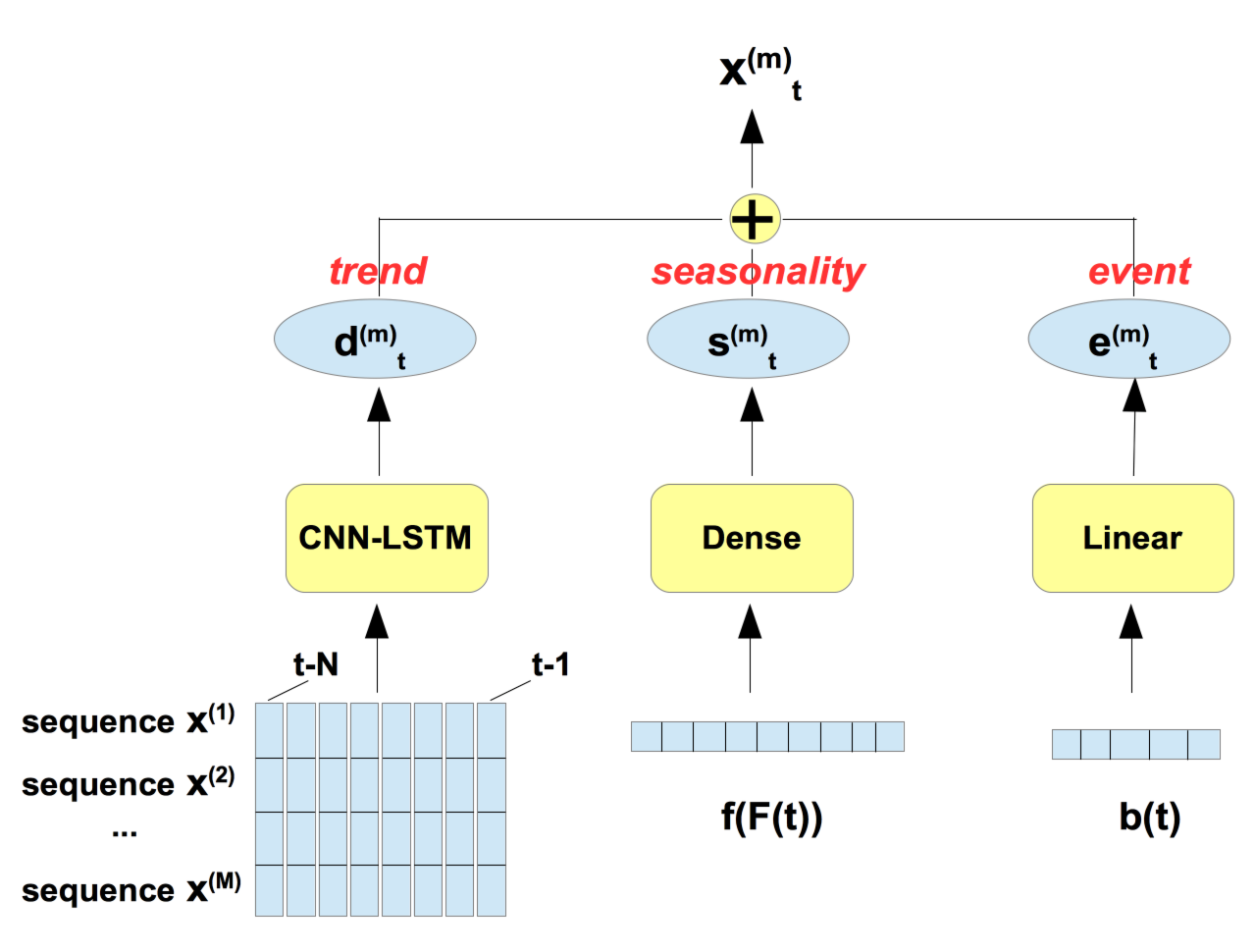}\\
	\caption{Architecture of our model.}
	\label{fig:stm}
	\vspace{-12pt}
\end{figure}

\subsection{Problem Formulation}

Multivariate time series are often encountered in many applications, and in many cases there exists strong correlation among multiple temporal sequences. For instance, running a machine learning task may involve the use of Amazon EC2, S3, and  relational database services (RDS). It is therefore expected the billings for these services are correlated. It is desirable if correlation among time series of different types is considered. Unlike the STM model, which assumes uni-variate time series input, we proposed a Deep Multivariate Structural Time series Model, which generalizes STM to handle multiple time series. We assume the input multivariate time series are $\{\bm{x}^{(1)},\bm{x}^{(2)},$$\dots$$ \bm{x}^{(M)}\}$, where $M$ is the number of input time series. We are interested in forecasting the $m$-th time series by rewriting Equation (\ref{eq:stm}) as
\begin{equation}
\vspace{-6pt}
\bm{x}_t^{(m)} = \bm{d}_t^{(m)} + \bm{s}_t^{(m)} +\bm{e}_t^{(m)}
\label{eq:mstm}
\vspace{0pt}
\end{equation}

In our model, the three components $\bm{d}_t^{(m)}$, $\bm{s}_t^{(m)}$, and $\bm{e}_t^{(m)}$ are learned jointly. The trend component $\bm{d}_t^{(m)}$ is learned by a CNN-LSTM hierarchical deep architecture, which takes multiple temporal sequences as the input, and is capable of leveraging correlations among the input temporal sequences and first-order temporal trending information. The seasonality and event components are learned by a fully connected neural net and linear model, respectively. The architecture of our model is demonstrated in Figure~\ref{fig:stm}. 

\subsection{The Trend Component}\label{subsec:trend}

The trend is the component of a time series that represents variations of low frequency in a time series. Predicting the trend is very important in many real-world applications, such as sales analysis, budget planning, and stock investment. In our model, we employ a CNN-LSTM hierarchical deep neural net to learn the trend of the time series. The CNN-LSTM neural net consists of two types of CNN layers, a one dimensional CNN layer and a two dimensional temporal CNN, followed by a LSTM layer. The hierarchical model enjoys several attractive properties. It is capable of (1) leveraging the correlations among multiple time series,  (2) capturing the trend of the time series by convolution of signals temporally, and (3) memorizing changes of trend over long sequences. In the following part, we will describe the input and output, the CNN layers, and the LSTM layer of our hierarchical neural net.

\textbf{\textit{(1) Input and output:}} The input for our trend learner includes $M$ time series. In addition, one important goal for extracting the trend component is to smooth out the irregular roughness to uncover a clearer signal. Many time series analysis models often only look at one-step lagged historical observation, which will introduce randomness in the forecasts of trend. To reduce the randomness in one lagged sample, our trend learner takes $M$ time series $\{\bm{x}^{(1)}_{t-N:t-1}, \bm{x}^{(2)}_{t-N:t-1}, \dots, \bm{x}^{(M)}_{t-N:t-1}\}$ at time $t-1$, each with $N (N>1)$ step lagged observations as the input. The input are then fed into CNN-LSTM hierarchical layer. The output is the estimation of the trend component for the time series that we would like to forecast.

\textbf{\textit{(2) 1D CNN layer:}} The dependency among multiple time series is leveraged using a 1D CNN layer consisting of $K_1$ kernels of size $M$. The multiple input time series $\{\bm{x}^{(1)}_{t-N:t-1}, \bm{x}^{(2)}_{t-N:t-1}, \dots, \bm{x}^{(M)}_{t-N:t-1}\}$ can be reshaped as a $M\times N$ matrix. The convolution is performed on each column of the matrix. The output of 1D CNN layer is a matrix $\mathbf{C}^{(1)}\in \mathbb{R}^{K_1\times N}$. Assume $\bm{w}^{k,1d}\in \mathbb{R}^{M}$ is the $k$-th kernel learned by the model. If we pick the first entry in the $k$-th row of $\mathbf{C}^{(1)}$, it can be expressed by
\begin{equation}
C_{k,1}^{(1)} = w^{k,1d}_1\cdot x^{(1)}_{t-N}+w^{k,1d}_2\cdot x^{(2)}_{t-N}+\dots+w^{k,1d}_M\cdot x^{(M)}_{t-N}
\label{eq:1d}
\end{equation}
From Equation \ref{eq:1d}, we can find the output of 1D CNN layer is equal to a weighted summation of the multiple input temporal sequences. By learning the weighting coefficients, we expect leveraging signals from all correlated time series can improve the forecasts of the time series we are interested in.

\textbf{\textit{(3) 2D temporal CNN layer:}} Differencing ~\cite{ham1994,tsay2005} is a technique commonly used in many time series models for extract the trend. For instance, ARIMA model ~\cite{box1970,hipel1994,cochrane1997,zhang2003} usually uses differencing to remove the temporal trend to produce more stationary time series. Take the first-order differencing as an example, it employs the transformation
\begin{equation}
\vspace{-6pt}
\text{diff} (x_t) = x_t-x_{t-1}
\label{eq:dif}
\vspace{0pt}
\end{equation}

The 2D temporal CNN layer employed by our model is somewhat similar in spirit to differencing. We would like to extract a \emph{weighted} trend/differencing of current signals for better forecasting of the future trend. Our 2D temporal CNN layer consists of $K_2$ kernels of size $M\times 2$, and the convolution is performed on both columns and rows of the input matrix. The output of this layer is a matrix $\mathbf{C}^{(2)}\in \mathbb{R}^{K_2\times(N-1)}$. Assume $\textbf{W}^{k,2d}\in \mathbb{R}^{M\times 2}$ is the $k$-th kernel learned by the model, then the first entry in the $k$-th row of $\textbf{C}^{(2)}$ can be expressed by
\begin{equation}
C_{k,1}^{(2)} = \sum_{m=1}^{M}\bigg(W^{k,1d}_{m,1}\cdot x^{(m)}_{t-N}+W^{k,1d}_{m,2}\cdot x^{(m)}_{t-N+1}\bigg)
\label{eq:2d}
\end{equation}
where in the right side $W^{k,1d}_{m,1}\cdot x^{(m)}_{t-N}+W^{k,1d}_{m,2}\cdot x^{(m)}_{t-N+1}$ is somewhat similar to differencing. However, the original differencing assumes fixed weighting coefficients for lagged observations (e.g. $1$ and $-1$ for $x_t$ and $x_{t-1}$ respectively in Equation \ref{eq:dif}), whereas the coefficients $W^{k,1d}_{m,1}$ and $W^{k,1d}_{m,2}$ in our model are learned.

\begin{figure}[h]
	\vspace{-12pt}
	\centering
	\includegraphics[scale = 0.35]{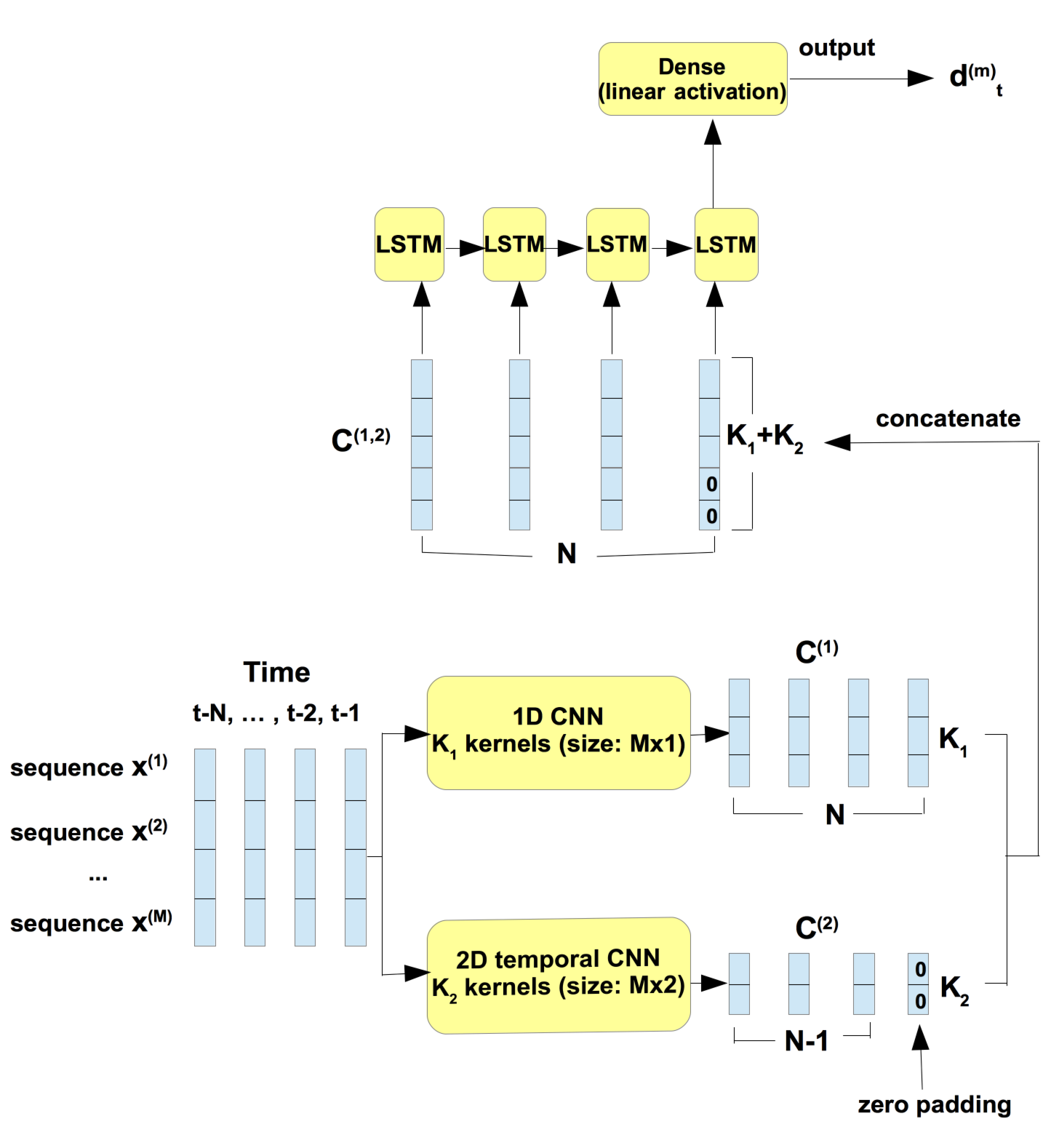}\\
	\caption{CNN-LSTM for learning the trend.}
	\label{fig:cnn-lstm}
	\vspace{-12pt}
\end{figure}

\textbf{\textit{(4) LSTM layer:}}  We use 1D and 2D CNN layers for feature extraction on multiple temporal sequences, and then stack with a long short-term memory (LSTM)~\cite{seep1997,patterson2017} layer followed by a Dense layer (with linear activation) to support temporal sequence forecast. The input for the LSTM layer is a matrix $\mathbf{C}^{(1,2)}$, which is obtained by row-wise concatenation of matrices $\mathbf{C}^{(1)}$ and $\mathbf{C}^{(2)}$. Note that $\mathbf{C}^{(1)}$ has $N$ columns, whereas $\mathbf{C}^{(2)}$ has $N-1$ columns. Zero padding is performed on the last column of $\mathbf{C}^{(2)}$ before concatenation. Details can be found in Figure \ref{fig:cnn-lstm}.
\subsection{Modelling The Seasonality Component}
Time series often shows cyclical patterns. For instance, sales in auto industry and real estate typically have changes of one year cyclical length. Fourier series~\cite{carslaw1921} has been used to approximate the cyclical behavior of the time series in existing works~\cite{harvey1993,fumi2013,sean2018}. In the study of Fourier series, complicated but periodic functions are written as the sum of simple waves mathematically represented by sines and cosines. Inspired by this, we model the seasonality component $\bm{s}_t$ of the time series as a nonlinear function of Fourier terms (pairs of sines and cosines of different frequencies).
\begin{equation}
\vspace{-6pt}
\bm{s}_t = f\big(F(t)\big)
\end{equation}
$F(t) = \big(\cos{\lambda(1)},\dots,\cos{\lambda(g)},\sin{\lambda(1)},\dots,\sin{\lambda(g)}\big)$ is the Fourier terms at time $t$, and $\lambda(g)=\frac{2\pi gt}{P}$. $P$ is the length of seasonal cycle (e.g. $P = 7$ for weekly seasonality, and $P=365$ for yearly seasonality), and $g$ is the number of Fourier terms, or the number of pre-defined frequencies. A smaller $g$ corresponds to applying a low-pass filter to the seasonality components, whereas a larger $g$ allows for fitting cyclical patters with more frequent changes. $f(\cdot)$ is a nonlinear function which is learned by a fully connected neural network.

Note that a time series may contain multiple seasonal cycles (e.g. weekly, yearly) of different lengths. In that case, multiple sets of Fourier terms with different seasonal cycle lengths can be used as the input for the fully connected neural net.

\subsection{Modeling The Event Component}

Specific events (e.g. holidays) can have significant impact on daily time series, causing strong upward or downward changes. For instance, retailers often see boosted sales during Black Friday and Thanksgiving Day. It is difficult for the trend and seasonality components to capture these changes. However, the event effect is usually fixed and somewhat predictable. Therefore, we add an additional component $\bm{e}_t$ to model the fixed event effect. We encoded the events as a binary regressor $\bm{b}_t \in \{0, 1\}^L$, where $L$ is the number of unique event types. Unlike seasonality component, $\bm{e}_t$ is assumed to be a simple linear function of a regressor
\begin{equation}
\vspace{-6pt}
\bm{e}_t = \bm{a}\bm{b}_t
\vspace{-2pt}
\end{equation}
where $\bm{a}\in \mathbb{R}^L$. The reason linear function is used is due to the fact that in this work the time series we use do not have complex event types. For data with a variation of different event types, nonlinear function, such as fully connect neural network, can be used.
\section{Results}
We evaluate our model, both qualitatively
(forecasts of the trend, seasonality, and event components) and quantitatively (in its ability
to forecast future values), by performing experiments on two real-world data sets. One is a set of Amazon AWS daily billing data for different services, such as Amazon Elastic Compute Cloud (EC2), Simple Storage Service (S3), and Relational Database Service (RDS), in the first few months of 2018. The other dataset is the closing price for stocks of Verizon and T-mobile, both of which are telecommunications companies, from 2013 to 2017.
\subsection{Experimental Setup and Baselines}
For the Amazon AWS billing time series, weekly seasonality ($P=7$) is considered since we find weekly cyclical patterns in the billings for specific service, such as S3. For the stock price, yearly seasonality ($P=365$) is considered. We used a smaller neural network with 4 filters for both 1D CNN and 2D CNN for AWS billing data, since this data has relatively fewer samples, and a relatively larger network with 16 filters for 1D and 2D CNN for the stock data since it has more samples. For both datasets, the number of hidden units for our model's LSTM layer is set as 8. Our model is implemented in Python using the open source neural network library Keras. Mean absolute error between the ground truth signal and estimation is used as the objective function, and Adam algorithm is utilized for optimization. We compared our proposed method with the following four state-of-the-art baselines.

\textit{(1) Seasonal ARIMA:} Autoregressive integrated moving average (ARIMA) model is a popular and widely used statistical method for time series forecasting. The basic assumption behind the ARIMA model is that a univariate time series is a combination of autoregressive (AR) and moving average (MA) lags which capture the autocorrelation within the time series. In seasonal ARIMA, AR and MA terms predict future values of time series using data values and errors at times with lags that are multiples of cyclical length.

\textit{(2) Prophet:} Prophet ~\cite{sean2018} is a Bayesian nonlinear univariate generative model for time series forecasting which was proposed by Facebook in 2018. Like our method, Prophet is also a structural time series analysis method, which explicitly models the trend, seasonality, and event effects. The parameters for cyclical length and event date for the Prophet are set the same as our model.

\textit{(3) Univariate LSTM:} The univariate LSTM model consists of one LSTM layer followed by a dense layer with linear activation. The input for the univariate LSTM model is a single time series. The univariate LSTM considers the same number of lagged steps of historical observations for training as our model does.

\textit{(4) Multivariate LSTM:} The multivariate LSTM (MLSTM) model have the same neural net architecture as the univariate LSTM. Unlike univaraite LSTM, the input for the multivariate LSTM model are multiple temporal sequences.

\subsection{Forecasts of The Trend, Seasonality and Event}

We picked two AWS billing time series to evaluate the proposed model's capability of learning the trend, seasonality and event components. Multivariate input is not considered here. The input for our model is a single time series, and the output is a one-step short-term forecast.

\begin{figure}[h]
	\vspace{-12pt}
	\centering
	\includegraphics[scale = 0.18]{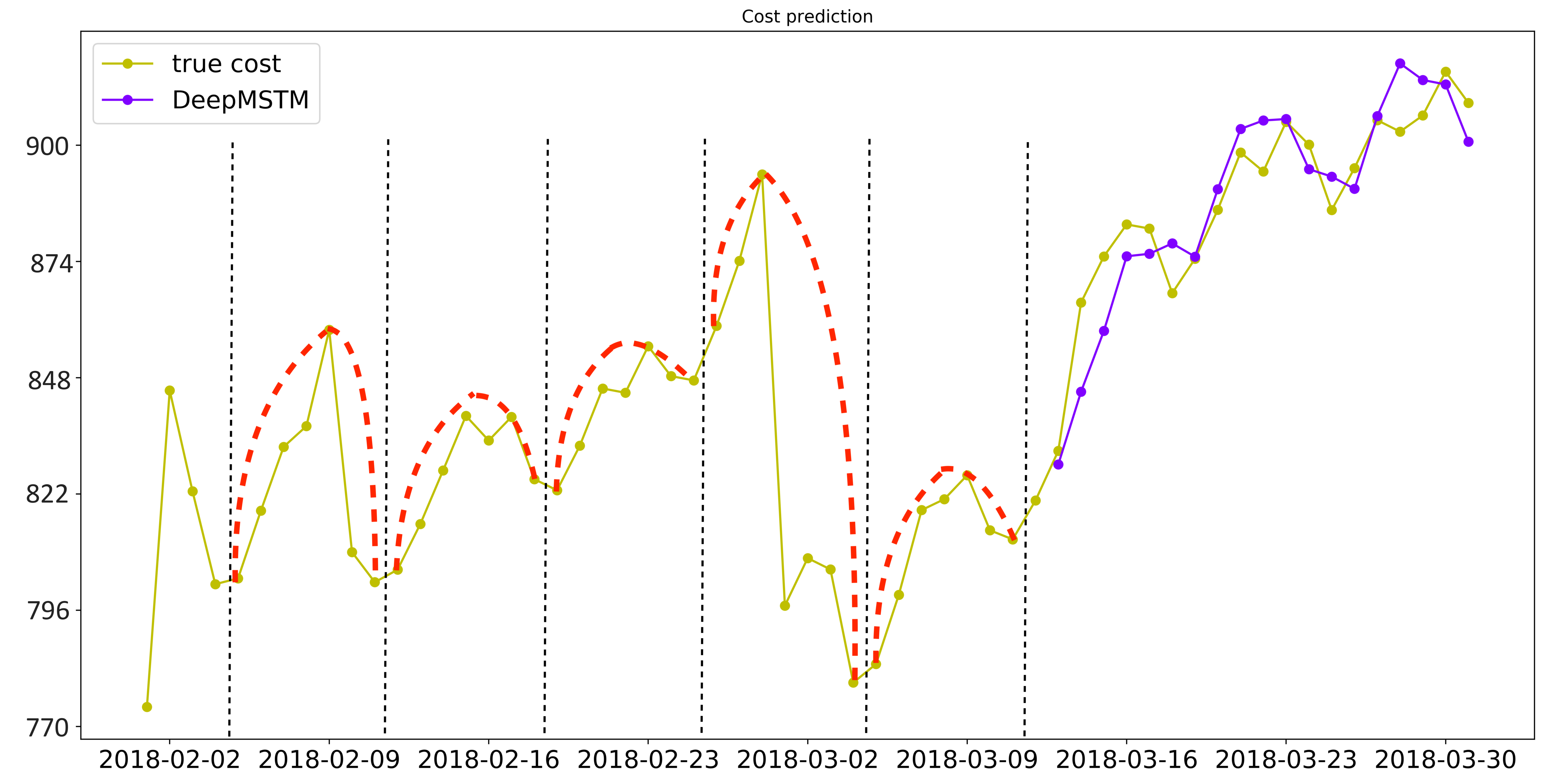}\\
	(a)\\
	\includegraphics[scale = 0.18]{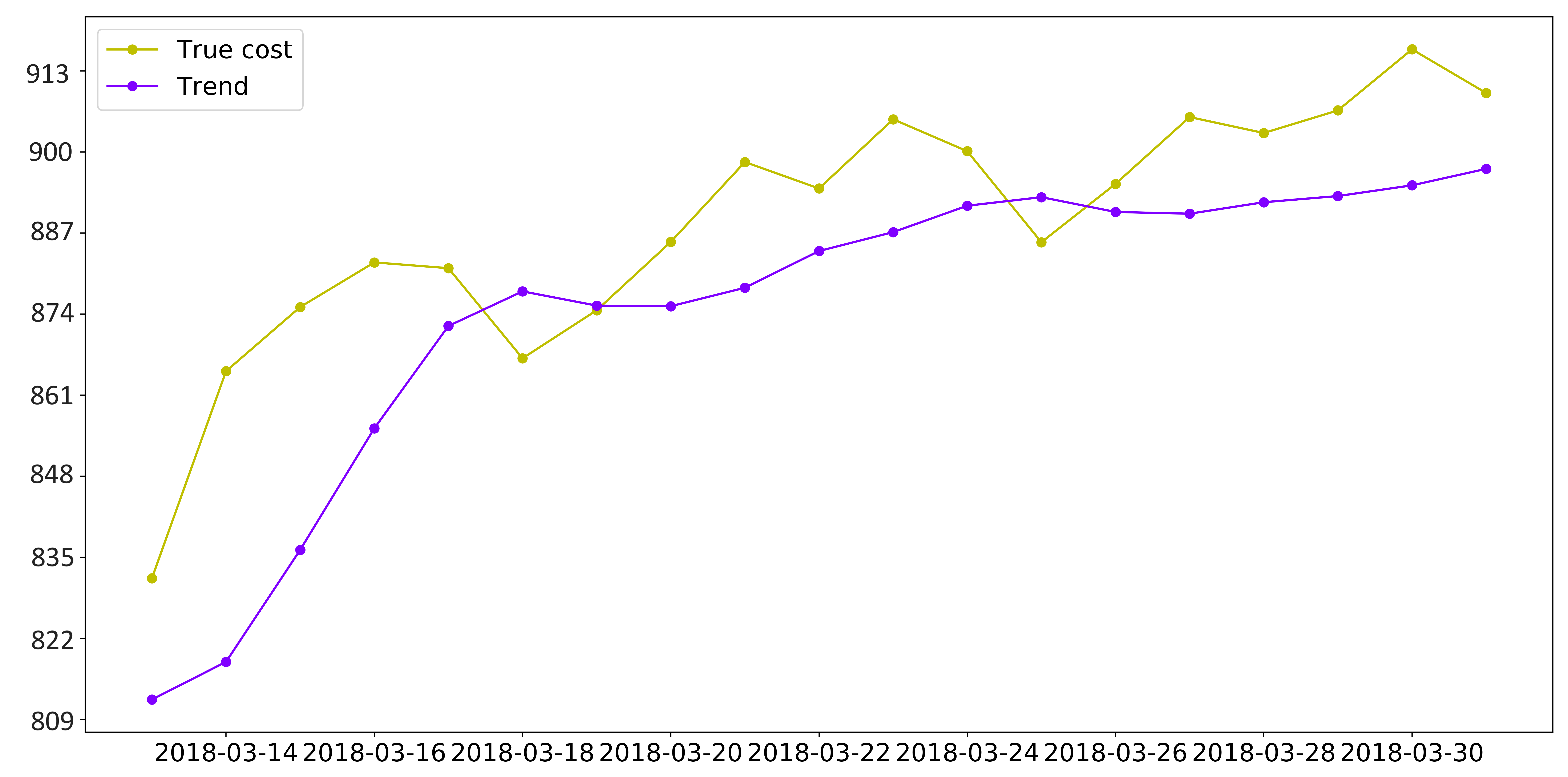}\\
	(b)\\
	\includegraphics[scale = 0.18]{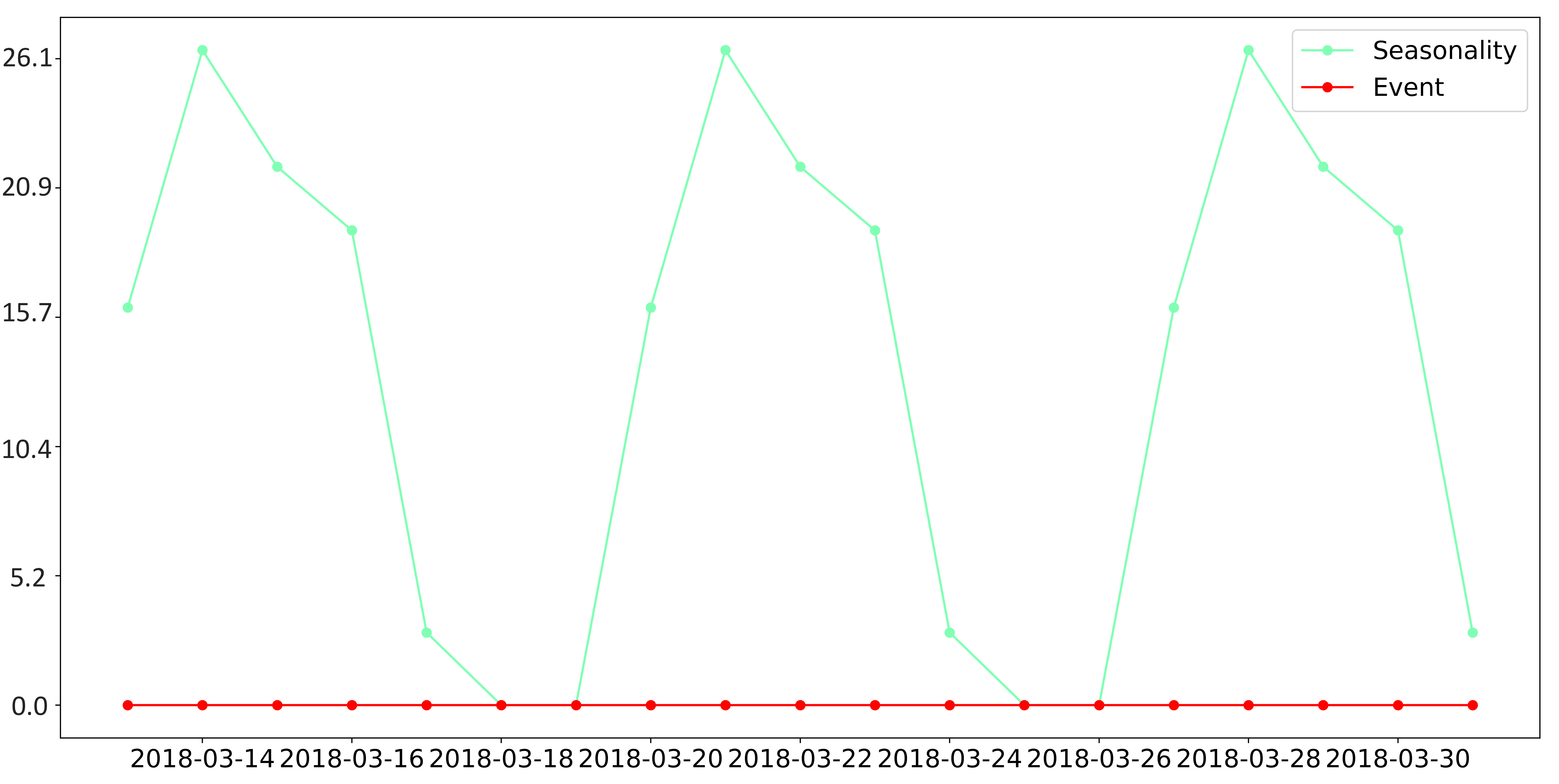}\\
	(c)\\
	\caption{AWS S3 billing forecast (a) and the zoomed true cost and predicted trend (b), seasonality, event components (c).}
	\label{fig:s3pred}
	\vspace{-12pt}
\end{figure}

The first time series is the S3 billing for an AWS account from February 2018 to March 2018, as shown in Figure \ref{fig:s3pred}(a). The data from February 1 to March 12 is used as the training set, and the rest is used for the testing. The time series has apparent changes in trend, and a weekly $\Lambda$-shaped upward-and-downward cyclical change, as indicated by the red dashed curve. The trend, seasonality, and event components for the forecasting in Figure \ref{fig:s3pred}(a) are demonstrated in \ref{fig:s3pred}(b) and (c). As observed, the learned trend component is consistent with the overall trend of the ground truth signal, and the $\Lambda$-shaped cyclical changes are successfully captured by the seasonality component.

\begin{figure}[h]
	\vspace{-12pt}
	\centering
	\includegraphics[scale = 0.18]{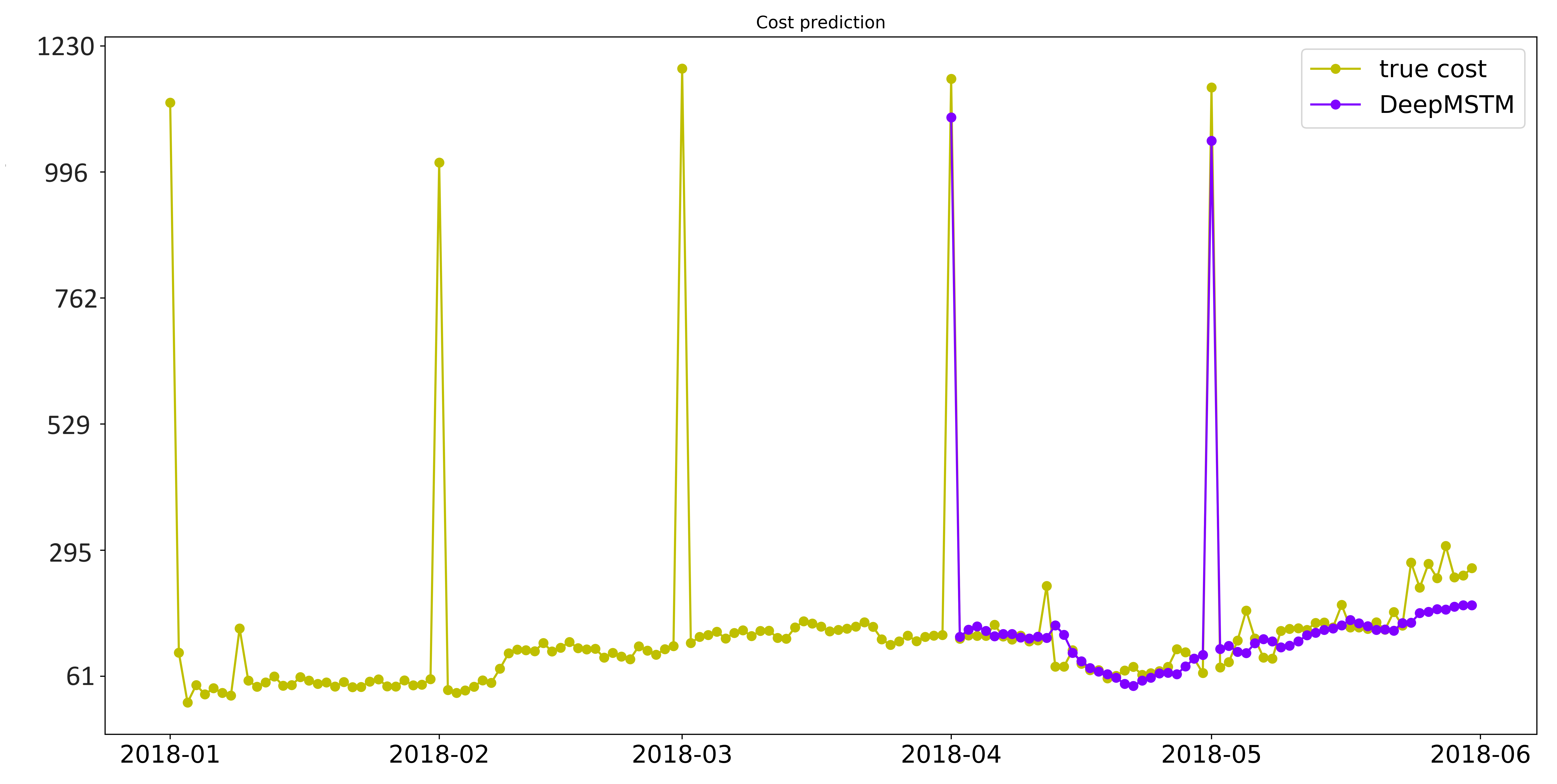}\\
	(a)\\
	\includegraphics[scale = 0.18]{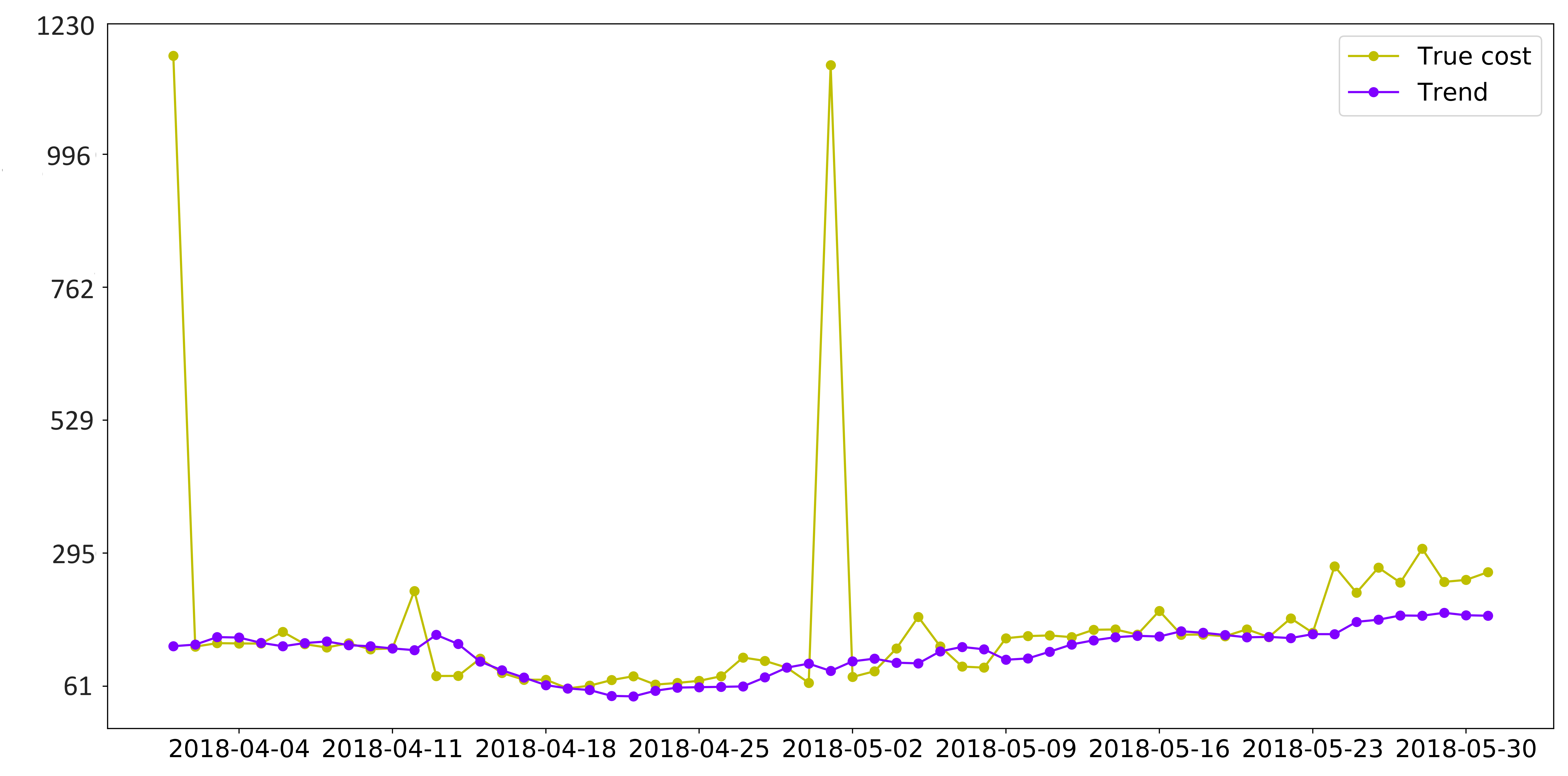}\\
	(b)\\
	\includegraphics[scale = 0.18]{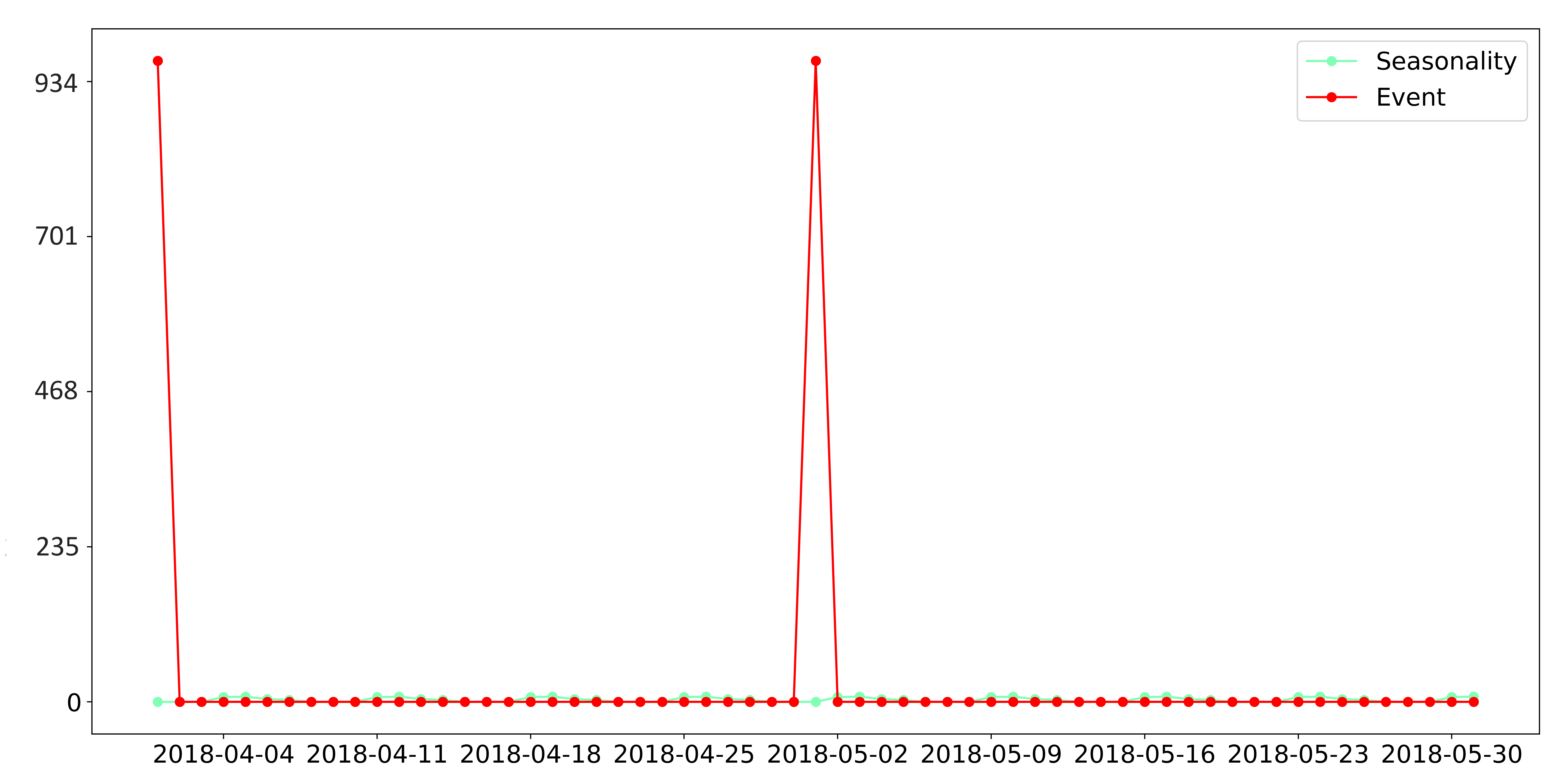}\\
	(c)\\
	\caption{AWS EC2 billing forecast (a) and the zoomed true cost and predicted trend (b), seasonality and event components (c).}
	\label{fig:ec2pred}
\end{figure}

The second time series we pick is the EC2 billing for the same AWS account from January 2018 to May 2018, as shown in Figure \ref{fig:ec2pred}(a). It has a relatively flat trend, but several spikes on the first day of each month. The spikes are caused by Reserved Instance Fee (RI Fee), which is usually charged on the first day of each month. We, therefore, set the first day of each month as the event date, and use a binary scalar $b(t)$ to encode whether it is an event date. Although here we only consider one type of event, more types of events can be easily encoded by using a binary vector $\mathbf{b}(t)\in\{0,1\}^L$ as we mentioned before. The trend, seasonality, and event components are shown in Figure \ref{fig:ec2pred}(b) and (c). As the figure shows, both the trend and seasonality components are relatively flat. However, the spiky signals caused by RI Fee is captured by the event component.

\subsection{Multivariate and Differencing}
To evaluate the benefits of introducing multivariate time series learning and the weighted ``differencing'' (using 2D CNN), we compare our model (Model 3), which includes 1D and 2D CNN and is trained using multivariate time series, with two variations. The first variation (Model 1) is trained on univariate input time series with only 2D CNN, and the second variation (Model 2) is trained on multivariate input time series but without 2D CNN (only consider 1D CNN). Two data sets are used in this experiment. The first one is the daily Amazon AWS billings for RDS and S3 for two AWS accounts from January 2018 to May 2018. For multivariate models, the inputs are RDS and S3 billing sequences, and the predicted time series is the S3 time series. Data from January to April are used for training, and the forecasts of billings for May are evaluated. The second data is the daily stock closing prices for Verizon (stock symbol: VZ) and T-mobile (stock symbol: TMUS) from 2013 to 2017. For multivariate models, the input are the closing prices of the two stocks, and then forecasts are performed on one stock each time. Data from January 2013 to April 2017 are used for training, and the remaining are used as testing data. Root mean square error (RMSE) is used to evaluate the performance. As the amplitudes for different time series are different, we also consider relative RMSE (RRMSE), which normalizes RMSE with the mean of the time series.
\begin{equation}
\text{RRMSE} = \text{RMSE}/(\frac{1}{N}\sum_{t=1}^{N}x_t)
\label{eq:rrmse}
\end{equation}
where $x_t$ is the ground truth value of the time series at time $t$. The results evaluated by RMSE and relative RMSE are shown in Table \ref{table:rmse-deep} and \ref{table:rrmse-deep} respectively. By comparing between Model 1 and Model 3, we find including correlated multivariate time series, such as AWS RDS billing or other stocks, leads to reduced errors for forecasts of S3 and stock closing prices. Furthermore, introducing the weighted differencing features extracted by 2D temporal CNN will result in more accurate trend forecasts, as indicated by the comparison between Model 2 and Model 3.

\begin{table}[h]
	\vspace{-12pt}
	\centering
	\caption{RMSE for the univariate DeepMSTM (second-order, denoted as model 1), multivariate DeepMSTM (first-order, denoted as model 2, and first-and-second-order, denoted as model 3).}
	\label{table:rmse-deep}
	{
		\begin{tabular}{|c|c|c|c|}
			\hline
			Data  & Model 1 & Model 2 & Model 3 \\
			\hline
			AWSID 1 & 11.740 & 10.563 & \textbf{9.275} \\
			AWSID 2 & 241.238 & 216.192 & \textbf{206.524}\\
			VZ & 8.001 & 2.200 & \textbf{2.168}\\
			TMUS& 4.651 & 2.850 & \textbf{2.730}\\
			\hline
		\end{tabular}
	}
\vspace{-12pt}
\end{table}

\begin{table}[h]
	\centering
	\caption{Relative RMSE for the univariate DeepMSTM (second-order, denoted as model 1), multivariate DeepMSTM (first-order, denoted as model 2, and first-and-second-order, denoted as model 3).}
	\label{table:rrmse-deep}
	{
		\begin{tabular}{|c|c|c|c|}
			\hline
			Data  & Model 1 & Model 2 & Model 3 \\
			\hline
			AWSID 1 & 0.061 & 0.055 & \textbf{0.048} \\
			AWSID 2 & 0.144 & 0.129 & \textbf{0.123}\\
			VZ & 0.173 & 0.048 & \textbf{0.047}\\
			TMUS & 0.074 & 0.045 & \textbf{0.043}\\
			\hline
		\end{tabular}
	\vspace{-12pt}
	}
\end{table}

\subsection{Comparing with the Baselines}

The comparison of our DeepMSTM model with Prophet, seasonal ARIMA, univariate LSTM and multivariate LSTM (MLSTM) are demonstrated in Table \ref{table:rmse} and \ref{table:rrmse}. The data sets and the training/testing data splits used in this experiment are the same as the last section. As can be observed, DeepMSTM consistently perform better than other baselines. In our experiments, we find multivariate LSTM always outperforms univariate LSTM, which further evidence the fact that including correlated multivariate time series can help. Furthermore, by comparing our DeepMSTM with multivariate LSTM model, we can conclude that the 1D CNN and 2D temporal CNN indeed lead to improved performance for the forecasts.

\begin{figure}[h]
	\vspace{-10pt}
	\centering
	\includegraphics[scale = 0.15]{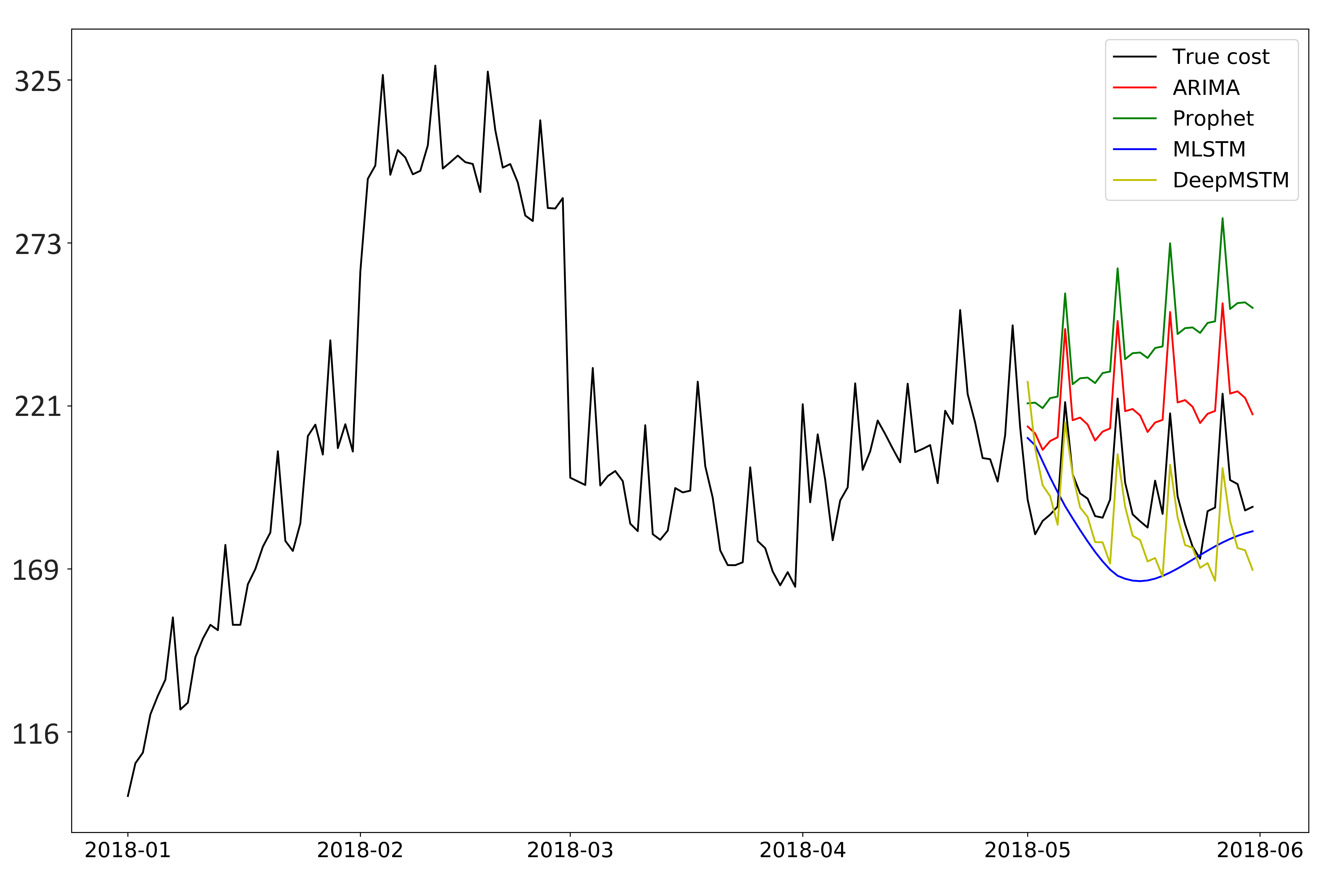}\\
	(a)\\
	\includegraphics[scale = 0.15]{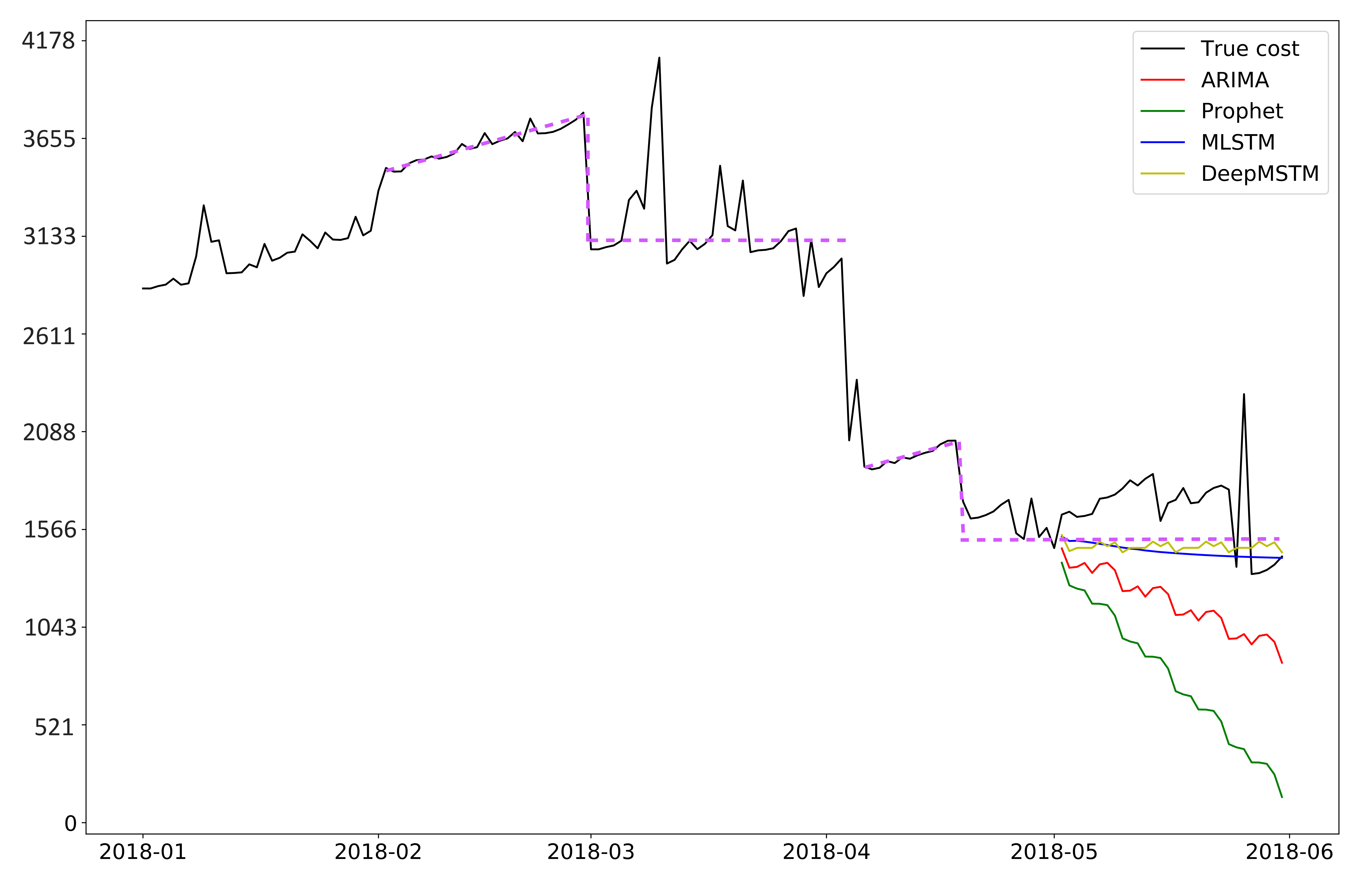}\\
	(b)\\
	\caption{AWS S3 billing forecast for ID 1 (a) and ID 2 (b) using seasonal ARIMA, Prophet, Multivariate LSTM, DeepMSTM.}
	\label{fig:baselines}
	\vspace{-11pt}
\end{figure}

The forecasts for the Amazon AWS S3 billings of the two AWSIDs obtained by Prophet, seasonal ARIMA, MLSTM, and DeepMSTM are shown in Figure \ref{fig:baselines}. Due to the relatively small error differences among all models for the stock data, we do not present the plot for stock data here. For AWSID 1, the S3 billing shows apparent weekly spikes, this seasonality change is successfully forecasted by seasonal ARIMA, Prophet, and DeepMSTM since all the three methods consider seasonality, whereas forecast by MLSTM is very smooth, and fails to learn the weekly changes. For AWSID 2, it is interesting to notice that both MLSTM and DeepMSTM seem to somehow remember the ``Z'' shaped changing pattern they observed far before, as indicated by the purple dotted lines. This is probably due to the LSTM layer's capability to ``memorize'' long-term patterns. while the forecasts for seasonal ARIMA and Prophet mainly depends on the trend of the most recent historical data. 

\begin{table}[h]
	\centering
	\caption{RMSE for the proposed model and baselines (ULSTM and MLSTM denotes univariate LSTM and multivariate LSTM respectively).}
	\label{table:rmse}
	{
		\begin{tabular}{|@{}c@{}|@{}c@{}|@{}c@{}|@{}c@{}|@{}c@{}|@{}c@{}|}
			\hline
			Data  & DeepMSTM & Prophet & ARIMA & ULSTM & MLSTM\\
			\hline
			AWSID 1 & \textbf{9.275} & 50.418 & 29.348 & 16.446 & 16.349\\
			AWSID 2 & \textbf{206.524} & 993.890 & 448.196 & 268.214 & 241.238\\
			VZ &\textbf{2.168}&3.031&2.586&15.036&2.169\\
			TMUS&\textbf{2.730}&14.515&4.910 & 5.550 & 3.105\\
			\hline
		\end{tabular}
	}\vspace{-10pt}
\end{table}

\begin{table}[h]
	\centering
	\caption{Relative RMSE for the proposed model and baselines (ULSTM and MLSTM denotes univariate LSTM and multivariate LSTM respectively).}
	\label{table:rrmse}
	{
		\begin{tabular}{|@{}c@{}|@{}c@{}|@{}c@{}|@{}c@{}|@{}c@{}|@{}c@{}|}
			\hline
			Data  & DeepMSTM & Prophet & ARIMA & ULSTM & MLSTM \\
			\hline
			AWSID 1 & \textbf{0.048} & 0.263 & 0.153 & 0.085 & 0.085\\
			AWSID 2 & \textbf{0.123} & 0.594 & 0.267 & 0.160 & 0.144\\
			VZ & \textbf{0.047} & 0.065 & 0.055 & 0.325 & 0.047\\
			TMUS& \textbf{0.043} & 0.230 & 0.077 & 0.088 & 0.049\\
			\hline
		\end{tabular}
	}
\vspace{-12pt}
\end{table}
\vspace{-8pt}
\section{Conclusion}
\label{summary}
We have presented the CNN-LSTM model for structural time series analysis which learns the trend, seasonality, and event components jointly. It allows data analysts to incorporate their knowledge/understanding about cyclical patterns and irregular changes related to specific events into the model. The extracted trend, seasonality, and event components can give interpretable results for further qualitative analyses. Our method enriches the family of time series analysis models by seamlessly leveraging the correlations among multivariate time series, as well as extracting the weighted differencing/trend feature, and leads to improved performance in time series forecasts. To the best of our knowledge, our method is the first framework that extends structural time series models to a deep architecture.
\vspace{-12pt}
\bibliography{ref}

\begin{thebibliography}{10}

\bibitem{box1970}
George Box and Jenkins Gwilym.
\newblock Time series analysis: Forecasting and control.
\newblock Holden-Day, San Francisco, CA, 1970.

\bibitem{kay2015}
Kay Brodersen, Fabian Gallusser, Jim Koehler, Nicolas Remy, and Steven Scott.
\newblock Inferring causal impact using bayesian structural time-series models.
\newblock In {\em Annals of Applied Statistics}, pages 247--274, 2015.

\bibitem{carslaw1921}
Horatio~Scott Carslaw.
\newblock Chapter 7: Fourier's series.
\newblock {\em Introduction to the Theory of Fourier's Series and Integrals},
  1:196, 1921.

\bibitem{cho2014properties}
Kyunghyun Cho, Bart Van~Merri{\"e}nboer, Dzmitry Bahdanau, and Yoshua Bengio.
\newblock On the properties of neural machine translation: Encoder-decoder
  approaches.
\newblock {\em arXiv preprint arXiv:1409.1259}, 2014.

\bibitem{cochrane1997}
John Cochrane.
\newblock Time series for macroeconomics and finance.
\newblock Graduate School of Business, University of Chicago, Spring, 1997.

\bibitem{donahue2015long}
Jeffrey Donahue, Lisa Anne~Hendricks, Sergio Guadarrama, Marcus Rohrbach,
  Subhashini Venugopalan, Kate Saenko, and Trevor Darrell.
\newblock Long-term recurrent convolutional networks for visual recognition and
  description.
\newblock In {\em Proceedings of the IEEE conference on computer vision and
  pattern recognition}, pages 2625--2634, 2015.

\bibitem{fumi2013}
Andrea Fumi, Arianna Pepe, Laura Scarabotti, and Massimiliano Schiraldi.
\newblock Fourier analysis for demand forecasting in fashion company.
\newblock In {\em International Journal of Engineering Business Management},
  2013.

\bibitem{patterson2017}
Adam Gibson and Josh Patterson.
\newblock Deep learning: A practitioner's approach.
\newblock In {\em O'Reilly Media}, 2017.

\bibitem{ham1994}
James~D. Hamilton.
\newblock Time series analysis.
\newblock Princeton University Press, 1994.

\bibitem{harvey1990}
Andrew Harvey and Simon Peters.
\newblock Estimation procedures for structural time series models.
\newblock In {\em Journal of Forecasting}, pages 89--108, 1990.

\bibitem{harvey1993}
Andrew Harvey and Neil Shephard.
\newblock Structural time series models.
\newblock In {\em Handbook of Statistics}, pages 261--302, 1997.

\bibitem{hipel1994}
Keith Hipel and Ian McLeod.
\newblock Time series modelling of water resources and environmental systems.
\newblock Amsterdam, Elsevier, 1994.

\bibitem{hochreiter1998vanishing}
Sepp Hochreiter.
\newblock The vanishing gradient problem during learning recurrent neural nets
  and problem solutions.
\newblock {\em International J of Uncertainty, Fuzziness and Knowledge-Based
  Systems}, 6(02):107--116, 1998.

\bibitem{hochreiter1997long}
Sepp Hochreiter and J{\"u}rgen Schmidhuber.
\newblock Long short-term memory.
\newblock {\em Neural computation}, 9(8):1735--1780, 1997.

\bibitem{seep1997}
Sepp Hochreiter and Jürgen Schmidhuber.
\newblock Long short-term memory.
\newblock In {\em Neural Computation}, pages 1735--1780, 1997.

\bibitem{kim2014}
Yoon Kim.
\newblock Convolutional neural networks for sentence classification.
\newblock In {\em Conference on Empirical Methods in NLP}, 2014.

\bibitem{kim2016character}
Yoon Kim, Yacine Jernite, David Sontag, and Alexander Rush.
\newblock Character-aware neural language models.
\newblock In {\em AAAI}, page 2741, 2016.

\bibitem{lahmiri2018minute}
Salim Lahmiri.
\newblock Minute-ahead stock price forecasting based on singular spectrum
  analysis and support vector regression.
\newblock {\em Applied Mathematics and Computation}, 320:444--451, 2018.

\bibitem{pai2005hybrid}
PingFeng Pai and ChihSheng Lin.
\newblock A hybrid arima and support vector machines model in stock price
  forecasting.
\newblock {\em Omega}, page 497, 2005.

\bibitem{sean2018}
Sean Taylor and Benjamin Letham.
\newblock Forecasting at scale.
\newblock In {\em The American Statistician}, pages 37--45, 2018.

\bibitem{tsay2005}
Ruey Tsay.
\newblock Analysis of financial time series.
\newblock John Wiley and SONS, 2005.

\bibitem{zhang2003}
Peter Zhang.
\newblock Time series forecasting using a hybrid arima and neural network
  model.
\newblock pages 159--175, 2003.

\end{thebibliography}
\bibliographystyle{plain}
\end{document}